\documentclass[twocolumn,a4paper]{paper}
\usepackage[sort&compress,square,comma,authoryear]{natbib}
\usepackage{float}
\usepackage[affil-it]{authblk}
\usepackage{amsmath}
\usepackage{mathtools}
\usepackage{rotating}

\usepackage{graphicx}
\usepackage{amssymb}
\usepackage{lineno}

\usepackage{commath}

\usepackage{url}

\usepackage{stfloats}

\usepackage[dvipsnames,table]{xcolor}
\definecolor{backcream}{HTML}{FFF0C8}
\usepackage{pdfpages}

\AddToShipoutPictureFG*{% Add picture to background of every page
	\AtPageUpperLeft{%
		\raisebox{-50pt}{\makebox[\paperwidth]{\noindent\fcolorbox{red}{backcream}{\begin{minipage}{18.5cm}\centering
						{\color{red}\textbf{This manuscript is now published in Frontiers in Computational Neuroscience.} \textbf{Please cite it as:}}\\
						Masquelier, T. and Kheradpisheh, S. R. (2018). Optimal localist and distributed coding of spatiotemporal spike patterns
						through STDP and coincidence detection. Frontiers in Computational Neuroscience, 12:74 (\url{https://doi.org/10.3389/fncom.2018.00074}).
		\end{minipage}}}}%
	}
}

\begin{document}
	
\title{\vspace{-3.0cm}Optimal localist and distributed coding of spatiotemporal spike patterns through STDP and coincidence detection}



%\author{Timoth\'ee Masquelier }
%\ead{timothee.masquelier@cnrs.fr}

%\affil{CERCO UMR5549 CNRS - Universit\'e Toulouse 3, France}

\author{Timoth\'ee Masquelier$^{1,2,}$%
	\thanks{e-mail: \texttt{timothee.masquelier@cnrs.fr}}}
\author{Saeed Reza Kheradpisheh$^{3}$}

\affil{\normalsize $^1$Centre de Recherche Cerveau et Cognition, UMR5549 CNRS - Universit\'e Toulouse 3, Toulouse, France.}
\affil{$^2$Instituto de Microelectr\'onica de Sevilla (IMSE-CNM), CSIC, Universidad de Sevilla, Sevilla, Spain.}
\affil{$^3$Department of Computer Science, Faculty of Mathematical Sciences and Computer, Kharazmi University, Tehran, Iran.}

\maketitle
	
%\hspace{13.9cm}1
%
%\ \vspace{20mm}\\
%
%{\LARGE Optimal localist and distributed coding of spatiotemporal spike patterns through STDP and coincidence detection}
%
%\ \\
%{\bf \large Timoth\'ee Masquelier$^{\displaystyle 1, \displaystyle 2}$, Saeed Reza Kheradpisheh$^{\displaystyle 3}$}\\
%{$^{\displaystyle 1}$CERCO UMR 5549, CNRS -- Universit\'e de Toulouse 3, F-31300, France.}\\
%{$^{\displaystyle 2}$IMSE (CNM),CSIC, Universidad de Sevilla, Spain.}\\
%{$^{\displaystyle 3}$Department of Computer Science, School of Mathematics, Statistics, and Computer Science, University of Tehran, Tehran, Iran.}\\
%
%%
%
%%\ \\[-2mm]
%{\bf Keywords:} neural coding, coincidence detection, leaky integrate-and-fire neuron, multi-neuron spike sequence, spatiotemporal spike pattern, unsupervised learning, spike-timing-dependent plasticity (STDP).
%
%\thispagestyle{empty}
%\markboth{}{NC instructions}
%%
%\ \vspace{-0mm}\\
%
%Abstract
\begin{center} {\bf Abstract} \end{center}
Repeating spatiotemporal spike patterns exist and carry information. Here we investigated how a single spiking neuron can optimally respond to one given pattern (localist coding), or to either one of several patterns (distributed coding, i.e. the neuron's response is ambiguous but the identity of the pattern could be inferred from the response of multiple neurons), but not to random inputs. To do so, we extended a theory developed in a previous paper~\citep{Masquelier2016a}, which was limited to localist coding. More specifically, we computed analytically the signal-to-noise ratio (SNR) of a multi-pattern-detector neuron, using a threshold-free leaky integrate-and-fire (LIF) neuron model with non-plastic unitary synapses and homogeneous Poisson inputs. Surprisingly, when increasing the number of patterns, the SNR decreases slowly, and remains acceptable for several tens of independent patterns.

In addition, we investigated whether spike-timing-dependent plasticity (STDP) could enable a neuron to reach the theoretical optimal SNR. To this aim, we simulated a LIF equipped with STDP, and repeatedly exposed it to multiple input spike patterns, embedded in equally dense Poisson spike trains. The LIF progressively became selective to every repeating pattern with no supervision, and stopped discharging during the Poisson spike trains. Furthermore, tuning certain STDP parameters, the resulting pattern detectors were optimal. Tens of independent patterns could be learned by a single neuron using a low adaptive threshold, in contrast with previous studies, in which higher thresholds led to localist coding only.

Taken together these results suggest that coincidence detection and STDP are powerful mechanisms, fully compatible with distributed coding. Yet we acknowledge that our theory is limited to single neurons, and thus also applies to feed-forward networks, but not to recurrent ones.
%%%%%%%%%%%


\section{Introduction}

In a neural network, either biological or artificial, two forms of coding can be used: localist or distributed. With localist coding, each neuron codes (i.e. maximally responds) for one and only one category of stimulus (or stimulus feature). As a result, the category of the stimulus (or the presence of a certain feature) can be inferred from the response of this sole neuron, ignoring the other neurons' responses. Conversely, with distributed coding each neuron responds to multiple stimulus categories (or features) in a similar way. Therefore the response of each neuron is ambiguous, and the category of the stimulus, or the presence of a certain feature, can only be inferred from the responses of multiple neurons. Thus the distinction between the two schemes is the number of different stimuli to which a given neuron responds -- not the number of neurons which respond to a given stimulus. Indeed, a localist network can have redundancy, and use multiple ``copies'' of each category specific neuron~\citep{Thorpe1989a,Bowers2009}.

Does the brain use localist or distributed coding? This question has been, and still is, intensively debated. In practice, discriminating between the two schemes from electrophysiological recordings is tricky~\citep{QuianQuiroga2010}, since the set of tested stimuli is always limited, the responses are noisy, the thresholds are arbitrary and the boundaries between categories are fuzzy. Here we do not attempt to do a complete review of the experimental literature; but rather to summarize it. It is commonly believed that distributed coding is prevalent~\citep{Rolls1997,OReilly1998,Hung2005,Quiroga2008}, but there is also evidence for localist coding, at least for familiar stimuli, reviewed in~\citep{Bowers2009,Thorpe2009,Thorpe2011,Bowers2017a,Roy2017}.

The question of localist \emph{vs.} distributed coding is also relevant for artificial neural networks, and in particular for the recently popular deep neural networks. Most of the time, these networks are trained in a supervised manner, using the backpropagation algorithm~\citep{LeCun2015}. The last layer contains exactly one neuron per category, and backpropagation forces each neuron to respond more strongly when the stimulus belongs to the neuron's category. In other words, localist coding is imposed in the last layer. Conversely, the hidden layers are free to choose their coding scheme, which is supposedly optimal for the categorization task at hand. It is thus very interesting to analyze the chosen coding scheme. It is not easy to do such analysis on the brain (as explained above), but we can do it rigorously for computational models by computing the responses to huge amounts of images, and even synthesizing images that maximize the responses. Results indicate that some hidden neurons respond to one object category only~\citep{Zhou2015,Nguyen2016,Olah2017}, while others respond to multiple different objects~\citep{Nguyen2016,Olah2017}. Thus it appears that both localist and distributed codes can be optimal, depending on the task, the layer number, and the network parameters (number of layers, neurons, etc.).

Let us come back to the brain, in which computation is presumably implemented by spiking neurons performing coincidence detection~\citep{Abeles1982,Konig1996,Brette2015}. This observation raises an important question, which we tried to address in this theoretical paper: can coincidence detector neurons implement both localist and distributed codes? In this context, different stimuli correspond to different spatiotemporal input spike patterns. Here each pattern was generated randomly, leading to chance-level overlap between patterns. In addition, each pattern was jittered at each presentation, resulting in categories of similar, yet different, stimuli. Can a neuron respond to one, or several of these patterns, and not to random inputs? What is the required connectivity to do so in an optimal way? And finally, can this required connectivity emerge with spike-timing-dependent plasticity (STDP), in an unsupervised manner?

To address these questions, we extended a theory that we developed in a previous paper, but which was limited to one pattern only, i.e. localist coding~\citep{Masquelier2016a}, to the multi-pattern case. Briefly, we derived analytically the signal-to-noise ratio (SNR) of a multi-pattern detector, and investigated the conditions for its optimality.
In addition, using numerical simulations, we showed that a single neuron equipped with STDP can become selective to multiple repeating spike patterns, even without supervision
and that the resulting detectors can be close to the theoretical optimum. Surprisingly, a single neuron could robustly learn up to $\sim 40$ independent patterns (using parameters arguably in the biological range). This was not clear from previous simulations studies, in which neurons equipped with STDP only learned one pattern (localist coding)~\citep{Masquelier2008,Masquelier2009,Gilson2011,Humble2012,Hunzinger2012,Klampfl2013,Nessler2013,Kasabov2013a,Krunglevicius2015a,Sun2016,Masquelier2016a}, or two patterns~\citep{Yger2015}. This shows that STDP and coincidence detection are compatible with distributed coding.

\section{Formal description of the problem}

\begin{figure}[!htb]
	\centering
	\includegraphics[width=1\linewidth]{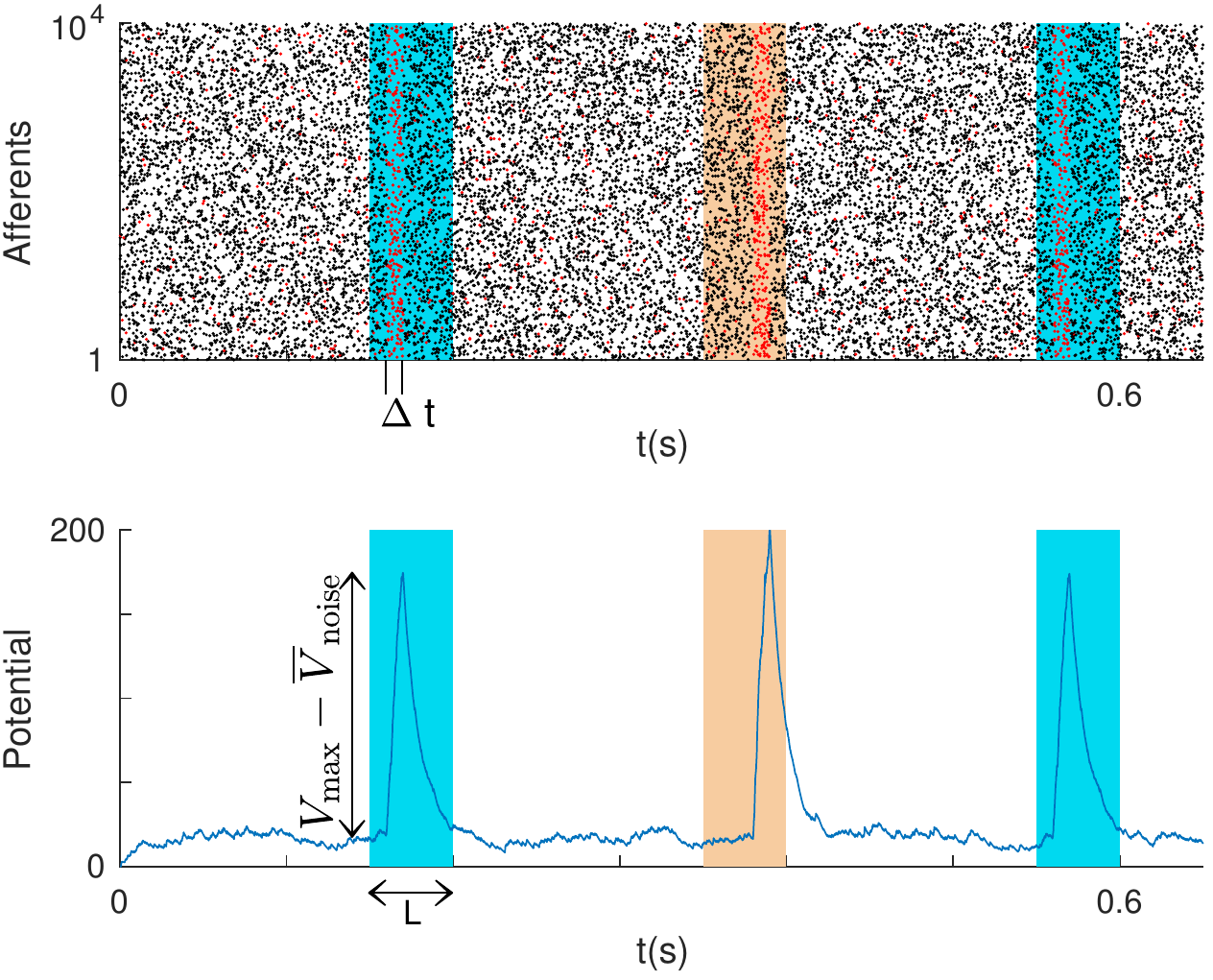}
	\caption{(Top) $P=2$ repeating spike patterns (colored rectangles) with duration $L$, embedded in Poisson noise. The LIF is connected to the neurons that fire in some subsections of the patterns with duration $\Delta t \leq L$ (these emit red spikes) (Bottom) The LIF potential peaks for patterns, and the double arrow indicates the peak height.}
	\label{fig:pb}
\end{figure}

The problem we addressed is similar to the one of~\citep{Masquelier2016a}, but extended to the multi-pattern case. For the reader's convenience, we fully describe it below.

We addressed the problem of detecting one or several spatiotemporal spike patterns with a single LIF neuron. Intuitively, one should connect the neurons that are active during the patterns (or during subsections of them) to the LIF neuron. That way, the LIF will tend to be more activated by the patterns than by some other inputs. More formally, we note $P$ the number of spike patterns, and assume that they all have the same duration  $L$. We note $N$ the number of neurons involved. For each pattern, we chose a subsection with duration  $\Delta t \le L$, and we connect the LIF to the $M$ neurons that emit at least one spike during at least one of these subsections (Fig.~\ref{fig:pb}).

We hypothesize that all afferent neurons fire according to a homogeneous Poisson process with rate $f$, both inside and outside the patterns. That is the patterns correspond to some realizations of the Poisson process, which can be repeated (this is sometimes referred to a ``frozen noise''). At each repetition a random time lag (jitter) is added to each spike, drawn from a uniform distribution over $[-T,T]$ (a normal distribution is more often used, but it would not allow analytical treatment~\citep{Masquelier2016a}).

We also assume that synapses are instantaneous, which facilitates the analytic calculations.

For now we ignore the LIF threshold, and we want to optimize its signal-to-noise ratio (SNR), defined as:
\begin{equation}
\label{eq:snr_def}
SNR =  \frac{V_{\mathrm{max}}-\overline{V}_{\mathrm{noise}}} {\sigma_{\mathrm{noise}}},
\end{equation}
where $V_{\mathrm{max}}$ is the maximal potential reached during the pattern presentations,  $\overline{V}_{\mathrm{noise}}$ is the mean value for the potential with Poisson input (noise period), and $\sigma_{\mathrm{noise}}$ is its standard deviation. Obviously, a higher $SNR$ means a larger difference between the LIF membrane potential during the noise periods and its maximum value, which occurs during the selected $\Delta t$ window of each pattern. Therefore, the higher the $SNR$ the lower the probability of missing patterns, and of false alarms.

We consider that $P$, $L$, $N$, $f$ and $T$ are imposed variables, and that we have the freedom to choose $\Delta t \le L$ and the membrane time constant $\tau$ in order to maximize the $SNR$.

We note that this problem is related to the synfire chain theory~\citep{Abeles1991}. A synfire chain consists of a series of pools of neurons linked together in a feed-forward chain, so that volleys of synchronous spikes can propagate from pool to pool in the chain. Each neuron can participate in several of such chains. The number of different chains that can coexist in a network of a given size has been termed capacity. This capacity can be optimized~\citep{Herrmann1995}. To do so, a given neuron should respond to certain spike volleys, but not to others, which is similar to our optimization of a multi-pattern $SNR$. Yet it is also different: we use homogeneous Poisson activity, not spike volleys, and we ignore the threshold, while synfire chains require thresholds.

\section{A theoretical optimum}
\subsection{Deriving the SNR analytically}

Here we are to find the optimum $SNR$ of the LIF for $P$ patterns. To this end we should first calculate the $SNR$ analytically.
Again, the derivations are similar to the ones in~\citep{Masquelier2016a}, but extended to the multi-pattern case (which turned to mainly impact Equation~\ref{eq:M}).

In this section, we assume non-plastic unitary synaptic weights. That is an afferent can be either connected ($w=1$) or disconnected ($w=0$) (in the Appendix we estimate the cost of this constraint on the $SNR$). Thus the LIF obeys the following differential equation:
\begin{equation}
\label{eq:LIF}
\tau\frac{\dif{V}}{\dif{t}}=-V+\tau\sum\limits_{i}\delta(t-t_i),
\end{equation}
\noindent where the $t_i$ are the presynaptic spike times of all the connected afferents.

Since synapses are instantaneous and firing is Poissonian, during the noise periods and outside the $\Delta t$ windows we have:
$\overline{V}_{\mathrm{noise}} = \tau f M$ and
$\sigma_{\mathrm{noise}} = \sqrt{\tau f M /2}$~\citep{Burkitt2006a}, where $M$ is the number of connected input neurons (with unitary weights).

To compute $V_{\mathrm{max}}$, it is  convenient to introduce the reduced variable:
\begin{equation}
\label{eq:v_max_def}
v_{\mathrm{max}}=\frac{V_{\mathrm{max}}-\overline{V}_{\mathrm{noise}}}{\overline{V}^{\infty}-\overline{V}_{\mathrm{noise}}},
\end{equation}
where  $\overline{V}^{\infty}=\tau r$ is the mean potential of the steady regime that would be reached if $\Delta t$ was infinite, and $r$ is the input spike rate during the $\Delta t$ window, resulting from the total received spikes from all input neurons during this window.

$v_{\mathrm{max}}$ can be calculated by exact integration of the LIF differential equation~\citep{Masquelier2016a}. Here we omit the derivation and present the final equation:
\begin{equation}
\begin{aligned}
&v_{\mathrm{max}} = \min \left( 1,\frac{\Delta t}{2T} \right) \\
&-\frac{\tau}{2T}\log \left(  1- e^{-\max(\Delta t, 2T)/\tau} + e^{-|\Delta t-2T|/\tau} \right).
\end{aligned}
\end{equation}

Using the definition of $v_{\mathrm{max}}$ in Equation~\ref{eq:v_max_def}, we can rewrite the $SNR$ equation as:
\begin{equation}
SNR =  v_{\mathrm{max}}\frac{\overline{V}^{\infty}-\overline{V}_{\mathrm{noise}}} {\sigma_{\mathrm{noise}}}.
\end{equation}

Obviously, different Poisson pattern realizations will lead to different values for $M$ and $r$ that consequently affect each of the terms $\overline{V}^{\infty}$, $\overline{V}_{\mathrm{noise}}$ and $\sigma_{\mathrm{noise}}$. Here we want to compute the expected value of the $SNR$ across different Poisson pattern realizations:
\begin{equation}
\label{eq:approx}
\begin{aligned}
\left<SNR\right> & =  v_{\mathrm{max}} \left< \frac{\overline{V}^{\infty}-\overline{V}_{\mathrm{noise}}} {\sigma_{\mathrm{noise}}} \right> \\
				 & =  v_{\mathrm{max}} \sqrt{2 \tau /f } \left< \frac{r - fM} {\sqrt{M}} \right>  \\
				 & \approx  v_{\mathrm{max}} \sqrt{2 \tau /f } \; \frac{\left<r\right> - f \left<M\right>} {\sqrt{\left<M\right>}}. 
\end{aligned}
\end{equation}
In Section~\ref{sec:validation} we justify this last approximation through numerical simulations, and we also show that this average $SNR$ is not much different from  the $SNR$ of particular Poisson realizations.

The last step to compute $\left<SNR\right>$ in Equation~\ref{eq:approx} is to calculate $\left<M\right>$ and $\left<r\right>$. Since firing is Poissonian with rate $\lambda = f\Delta t$, the probability that a given afferent fires at least once in a given pattern subsection of length $\Delta t$ is $p=1-e^{-f\Delta t}$.
Here, we consider independent patterns, i.e. with chance-level overlap. Hence the probability that a given afferent fires at least once in at least one of the $P$ pattern subsection is $1-(1-p)^P$. Thus the number of selected afferents $M$ is on average:
\begin{equation}
\label{eq:M}
%\left<M\right>=N\sum_{i=1}^{P} {{P}\choose{i}} (-1)^{i+1} (1-e^{-f\Delta t})^{i}.
\left<M\right>=N\left(1-(1-p)^P\right)=N\left(1-e^{-Pf\Delta t}\right).
\end{equation}

Finally, the expected effective input spike rate during the $\Delta t$ window is the expected total number of spikes, $fN\Delta t$, divided by $\Delta t$, thus:
\begin{equation}
\label{eq:r}
\left<r\right>=fN.
\end{equation}

We note that the $SNR$ scales with $\sqrt{N}$. In the rest of this paper we used $N=10^4$ afferents, which is in the biological range.
%We now have everything we need to compute $\left<SNR\right>$ from Equation~\ref{eq:approx}.

\subsection{Numerical validations}\label{sec:validation}

\begin{figure*}[!htb]
	\centering\includegraphics[width=1.0\linewidth]{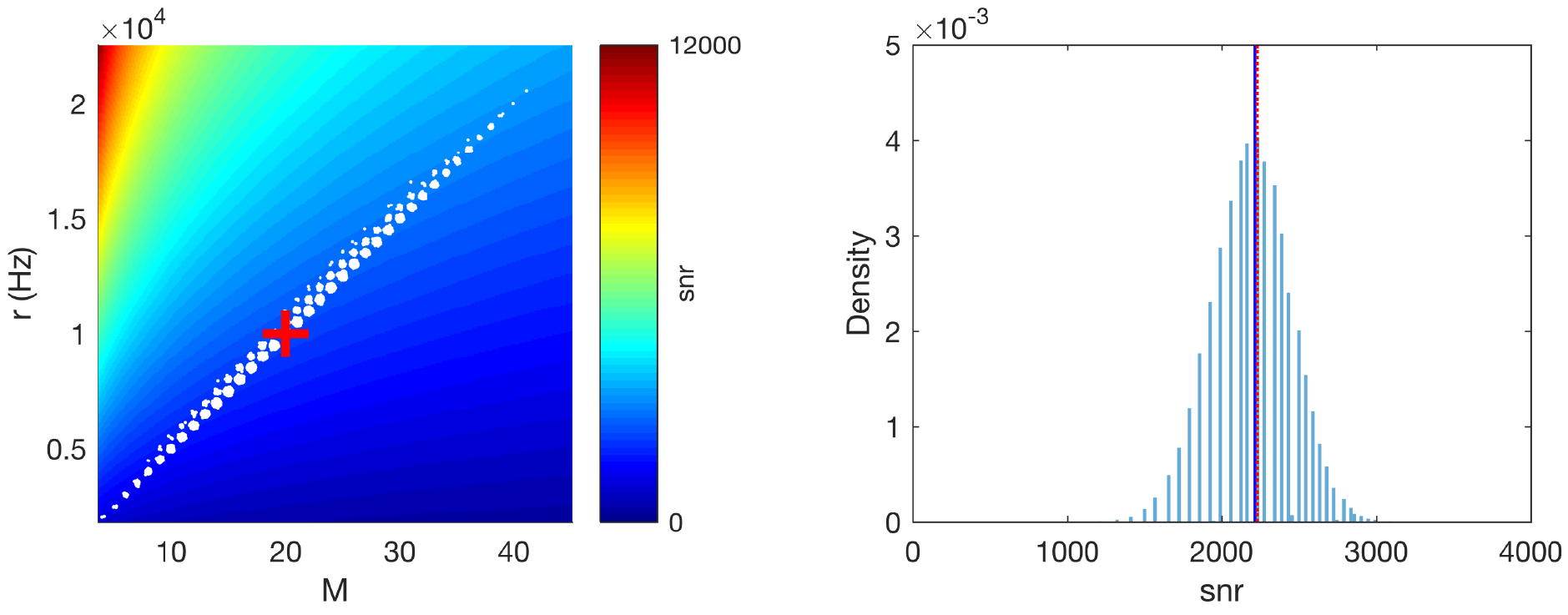}
	\caption{Numerical validation of the averaging operations. (Left) $M \times r$ plane. The white dots correspond to different realizations of a Poisson pattern (a jitter was added to better visualize density, given that both $M$ and $r$ are discrete). The background color shows the corresponding $snr$. The red cross corresponds to the average-case scenario $M = \left<M\right>$ and $r = \left<r\right>$. (Right) The distribution of $snr$ values across Poisson realizations. The  vertical blue solid line shows its average.
		The vertical red dotted line shows our approximation, $(\left<r\right> - f \left<M\right>)/\sqrt{\left<M\right>}$, which matches very well the true average. Parameters: $P=1$, $\Delta t=2$ms, $f=1$Hz. 
	}
	\label{fig:averaging}
\end{figure*}

We first checked if the variability of the $SNR$ across Poisson realizations is small, and also if the approximation we made to compute the average $SNR$ in Equation~\ref{eq:approx} is reasonable. To this aim, we generated $10^5$ Poisson patterns, and computed  $M$, $r$ and the reduced $SNR$, $snr = ( \left<r\right> - f \left<M\right> ) / \sqrt{\left<M\right>}$, for each of them (i.e. the right factor of the $SNR$ in Equation~\ref{eq:approx}, which is the only one that depends on the Poisson realization). As can be seen on Figure~\ref{fig:averaging} left, $M$ and $r$ are strongly correlated, and the data points lie near a line which corresponds to nearly constant $snr$ values (see the colored background). In other words, the $snr$ does not change much for different Poisson pattern realizations and the average $snr$ well represents the $snr$ distribution even for the worst and best cases.

In addition, as can be seen on Figure~\ref{fig:averaging} right, the average $snr$ across different Poisson patterns is very close to the $snr$ corresponding to the average-case scenario, i.e. $M = \left<M\right>$ and $r = \left<r\right>$ (as defined by Equations~\ref{eq:M} and~\ref{eq:r} respectively). Note that this Figure was done with relatively small values for the parameters $P$, $\Delta t$ and $f$ (respectively 1, 2ms, and 1Hz). Our simulations indicate that when increasing these parameter values, the approximation becomes even better (data not shown). 
 
\begin{figure}[!htb]
	\centering\includegraphics[width=.75\linewidth]{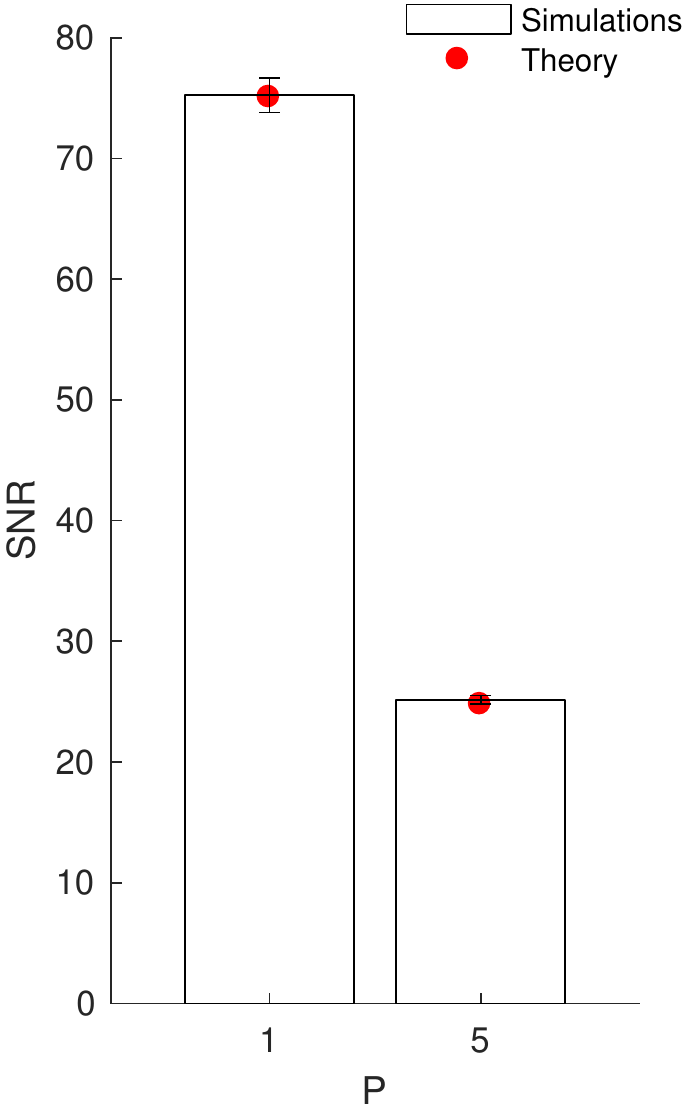}
	\caption{Numerical validation of the theoretical $SNR$ values, for $P=1$  and 5 patterns. Error bars show $\pm$1 s.d.
	}
	\label{fig:numerical}
\end{figure}

Next, we verified the complete $SNR$ formula (Eq.~\ref{eq:approx}), which also includes $v_{\mathrm{max}}$, through numerical simulations. We used a clock-based approach, and integrated the LIF equation using the forward Euler method with a 0.1ms time bin. We used $P=1$ and $P=5$ patterns, and performed 100 simulations with different random Poisson patterns of duration $L=20$ms with rate $f=5$Hz. We chose
$\Delta t=L=20$ms, i.e. the LIF was connected to all the afferents that emitted at least once during one of the patterns. In order to estimate $V_{\mathrm{max}}$, each pattern was presented 1000 times, every 400ms. The maximal jitter was $T=5$ms. Between pattern presentations, the afferents fired according to a Poisson process, still with rate $f=5$Hz, which allowed to estimate $\overline{V}_{\mathrm{noise}}$ and $\sigma_{\mathrm{noise}}$. We could thus compute the $SNR$ from Equation~\ref{eq:snr_def} (and its standard deviation across the 100 simulations), which, as can be seen on Figure~\ref{fig:numerical}, matches very well the theoretical values, for $P=1$ and 5. Note that the $SNR$ standard deviation is small, which confirms that the average $SNR$, i.e $\left<SNR\right>$, represents well the individual ones.

\subsection{Optimizing the SNR}

We now want to optimize the $SNR$ given by Equation~\ref{eq:approx}, by tuning $\tau$ and $\Delta t$. We also add the constraint $\tau fM\geq 10$ (large number of synaptic inputs), so that the distribution of V is approximately Gaussian~\citep{Burkitt2006a}. Otherwise, it would be positively skewed\footnote{With a low number of synaptic inputs, the mean V is close to zero. Since V is non-negative, its distribution is not symmetric anymore, but positively skewed.}, thus a high $SNR$ would not guarantee a low false alarm rate. We assume that $L$ is sufficiently large so that an upper bound for $\Delta t$ is not needed. We used the Matlab R2017a Optimization Toolbox (MathWorks Inc., Natick, MA, USA) to compute the optimum numerically.

\begin{figure*}[!htb]
	\centering
	\includegraphics[width=0.95\linewidth]{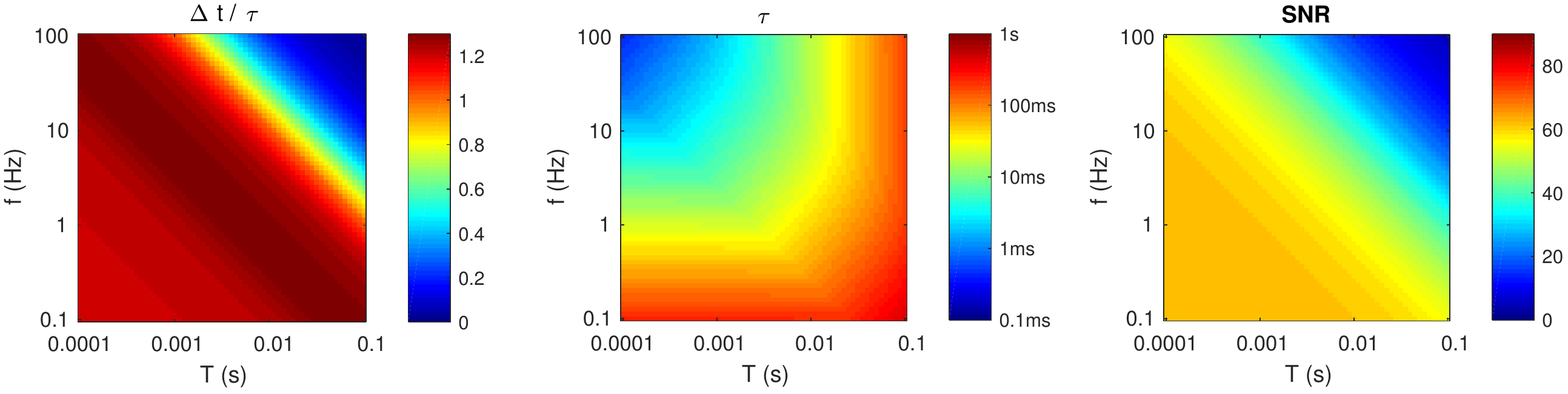}
	\caption{Optimal parameters for $P=2$, as a function of $f$ and $T$. (Left) Optimal $\Delta t$, divided by $\tau$. (Middle) Optimal $\tau$ (note the logarithmic colormap). (Right) Resulting $SNR$.}
	\label{fig:optim}
\end{figure*}

Figure~\ref{fig:optim} illustrates the results with $P=2$. One can make the following observations (similar to our previous paper which was limited to $P=1$ ~\citep{Masquelier2016a}):
\begin{itemize}
	\item Unless $f$ and $T$ are both high, the optimal $\tau$ and $\Delta t$  have the same order of magnitude (see Figure~\ref{fig:optim} left). 
	\item Unless $T$ is high ($>$10ms), or $f$ is low ($<$1Hz), then these timescales should be relatively small (at most a few tens of ms; see Figure~\ref{fig:optim} middle). This means that even long patterns (hundreds of ms or more) are optimally detected by a coincidence detector working at a shorter timescale, and which thus ignores most of the patterns. One could have thought that using $\tau \sim L$, to integrate all the spikes from the pattern would be the best strategy. But a long $\tau$ also decreases the detector's temporal resolution, thus patterns and random inputs elicit more similar responses, decreasing the $SNR$. Hence there is a trade-off, and it turns out that it is often more optimal to have  $\tau<L$, that is to use subpatterns as signatures for the whole patterns.
	\item Unsurprisingly, the optimal $SNR$ decreases with $T$ (see Figure~\ref{fig:optim} right). What is less trivial, is that it also decreases with $f$. In other words, sparse activity is preferable. We will come back to this point in the discussion.
\end{itemize}

What is the biological range for $T$, which corresponds to the spike time precision? Millisecond precision in cortex has been reported~\citep{Kayser2010,Panzeri2010,Havenith2011}. We are aware that other studies found poorer precision, but this could be due to uncontrolled variable or the use of inappropriate reference times~\citep{Masquelier2013}.

\begin{figure}[!htb]
	\centering
	\includegraphics[width=1\linewidth]{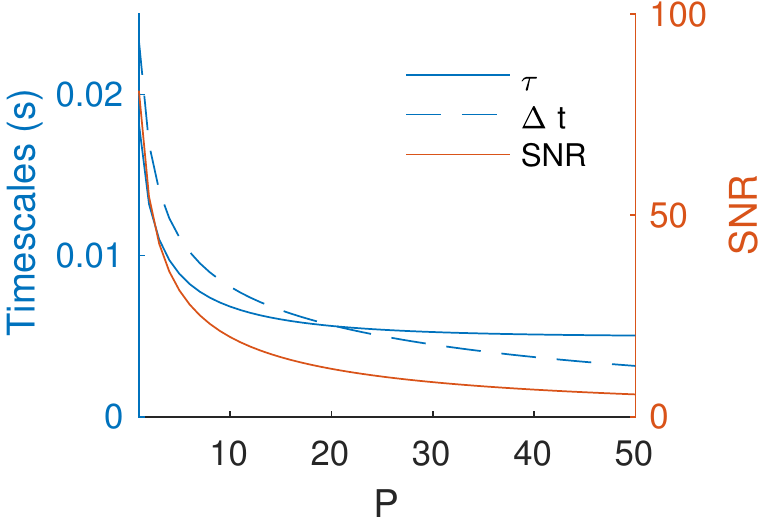}
	\caption{Optimal $\tau$ and $\Delta t$ (for $f=3.2$Hz, $T=3.2$ms) and resulting $SNR$ as a function of $P$.}
	\label{fig:varying_P}
\end{figure}

In the rest of the paper we focus,  as an example, on the point on the middle of the $T\times f$ plane -- $T=3.2$ms and $f=3.2$Hz. When increasing $P$, the optimal $\tau$ and $\Delta t$ decrease (Fig. \ref{fig:varying_P}). Unsurprisingly, the resulting $SNR$ also decreases, but only slowly. It thus remains acceptable for several tens of independent patterns (e.g. $SNR\sim$ 7 for $P=40$).

\section{Simulations show that STDP can be close-to-optimal}

\begin{sidewaysfigure*}
	\centering\includegraphics[width=1.0\linewidth]{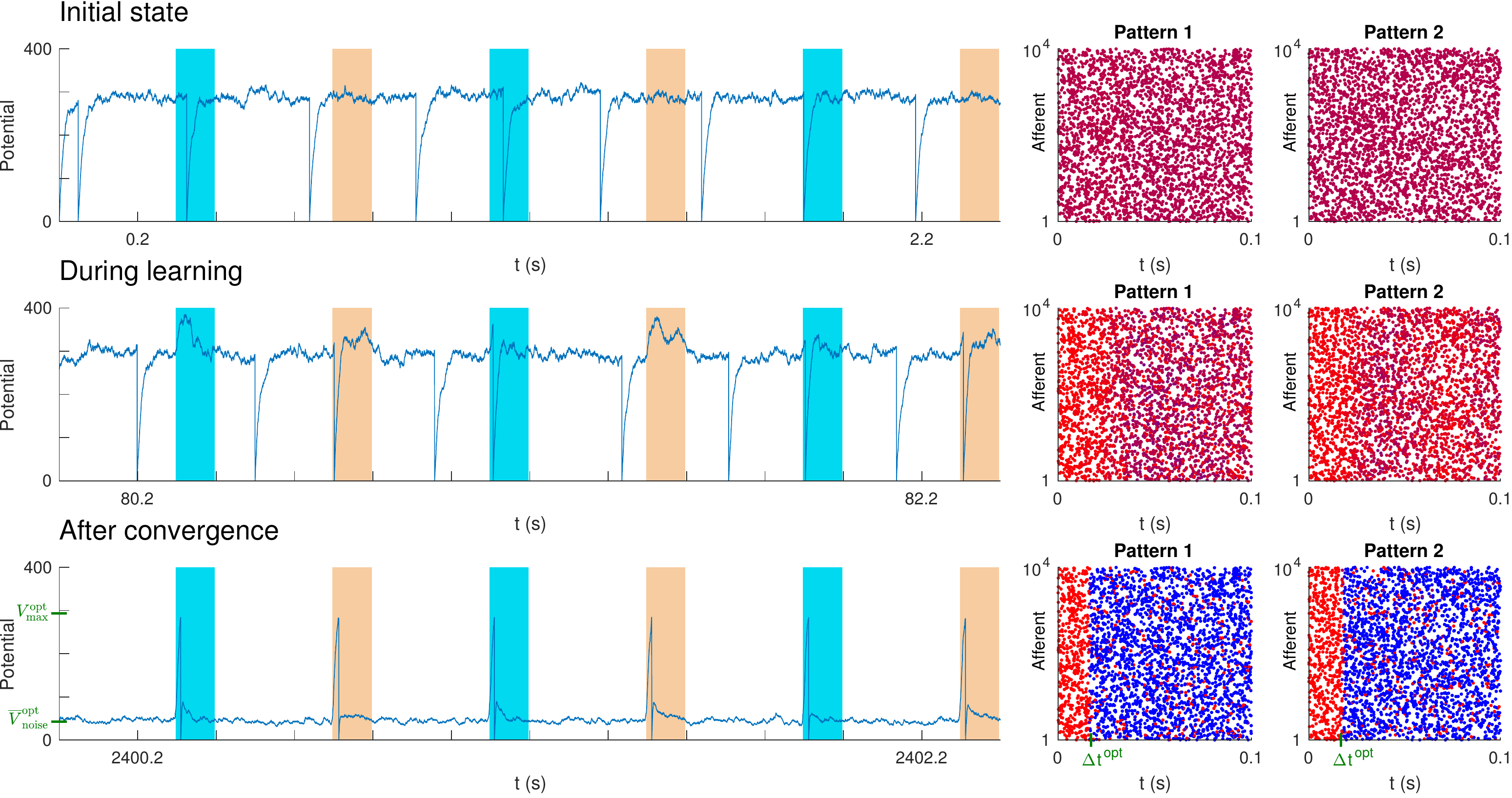}
	\caption{ Unsupervised STDP-based pattern learning. The neuron becomes selective to $P=2$ patterns. (Top) Initial state. On the left, we plotted the neuron's potential as a function of time. Colored rectangles indicate pattern presentations. Next, we plotted the two spike patterns, coloring the spikes as a function of the corresponding synaptic weights: blue for low weight (0), purple for intermediate weight, and red for high weight (1). Initial weights were uniform (here at 0.7, so the initial color is close to red). (Middle) During learning. Selectivity progressively emerges. (Bottom) After convergence. STDP has concentrated the weights on the afferents which fire at least once in at least one of the pattern subsections, located at the beginning of each pattern, and whose duration roughly matches the optimal $\Delta t$ (shown in green). This results in one postsynaptic spike  each time either one of the two pattern is presented. Elsewhere both $\overline{V}_{\mathrm{noise}}$ and $\sigma_{\mathrm{noise}}$ are low, so the $SNR$ is high. In addition $\overline{V}_{\mathrm{noise}}$ roughly matches the theoretical value $\overline{V}^{\mathrm{opt}}_{\mathrm{noise}}$ (shown in green), corresponding to the optimal $SNR$. We also show in green ${V}^{\mathrm{opt}}_{\mathrm{max}}$, the theoretical optimal value for ${V}_{\mathrm{max}}$. However, the potential never reaches it, because the adaptive threshold is reached before.
	}
	\label{fig:stdp}
\end{sidewaysfigure*}

Next we investigated, through numerical simulations, if STDP could turn a LIF neuron into an optimal multi-pattern detector. More specifically, since STDP does not adjust the membrane time constant $\tau$, we set it to the optimal value and investigated whether STDP could learn all the patterns with an optimal $\Delta t$\footnote{When $L$ is large (say tens of ms), STDP will typically not select all the afferents that fire in a full pattern, but only those that fire in a subsection of it, typically located at the beginning~\citep{Masquelier2008,Gilson2011,Masquelier2016a}, unless competition forces the neurons to learn subsequent subsections\citep{Masquelier2009}. The subsection duration depends on the parameters, and here we investigate the conditions under which this duration is optimal.}. Here, unlike in the previous section, we had to introduce a threshold, in order to have postsynaptic spikes, which are required for STDP. As a result, the optimal ${V}_{\mathrm{max}}$, computed in the previous section, was never reached. Yet a high ${V}_{\mathrm{max}}$ guarantees a low miss rate, and a low $\overline{V}_{\mathrm{noise}}$ guarantees a low false alarm rate. Optimizing the previously defined $SNR$ thus makes sense.

Again, we used a clock-based approach, and the forward Euler method with a 0.1ms time bin.
%The Matlab R2017a code for these simulations will be made available in ModelDB~\citep{Hines2004} at \texttt{https://senselab.med.yale.edu/modeldb/} once this paper is accepted in a peer-reviewed journal.
The Matlab R2017a code for these simulations has been made available in ModelDB~\citep{Hines2004} at \url{http://modeldb.yale.edu/244684}.

\subsection{Input spikes}

The setup we used was similar to the one of our previous studies~\citep{Masquelier2008,Masquelier2009,Gilson2011,Masquelier2016a}. Between pattern presentations, the input spikes were generated randomly with a homogeneous Poisson process with rate $f$. The $P$ spike patterns with duration $L=100$ms were generated only once using the same Poisson process (frozen noise). The pattern presentations occurred every $400$ms (in previous studies, we demonstrated that irregular intervals did not matter~\citep{Masquelier2008,Masquelier2009,Gilson2011}, so here regular intervals were used for simplicity). The $P$ patterns were presented alternatively, over and over again. Figure~\ref{fig:stdp} shows an example with $P=2$ patterns. At each pattern presentation, all the spike times were shifted independently by some random jitters uniformly distributed over $[-T,T]$.

\subsection{A LIF neuron with adaptive threshold}

We simulated a LIF neuron connected to all of the $N$ afferents with plastic synaptic weights $w_i\in[0,1]$, thus obeying the following differential equation:
\begin{equation}
\label{eq:LIF_w}
\tau\frac{\dif{V}}{\dif{t}}=-V+\tau\sum\limits_{i,j}w_i(t_{ij})\delta(t-t_{ij}),
\end{equation}
\noindent where $t_{ij}$ is the time of the $j^{\mathrm{th}}$ spike of afferent $i$.

We used an adaptive threshold (unlike in our previous studies~\citep{Masquelier2008,Masquelier2009,Gilson2011,Masquelier2016a}, in which a fixed threshold was used). This adaptive threshold was increased by a fixed amount ($1.8\theta_0$) at each postsynaptic spike, and then exponentially decayed towards its baseline value $\theta_0$ with a time constant $\tau_{\theta}=80$ms. This is a simple, yet good model of cortical cells, in the sense that it predicts very well the spikes elicited by a given input current~\citep{Gerstner2009,Kobayashi2009}. Here, such an adaptive threshold is crucial to encourage the neuron to learn multiple patterns, as opposed to fire multiple successive spikes to the same pattern. Since the theory developed in the previous sections ignored the LIF threshold, using an adaptive one is not worse than a fixed one, in the sense that it does not make the theory less valid.

We did not know which value for $\theta_0$ could lead to the optimum. We thus performed and exhaustive search, using a geometric progression with a ratio of 2.5\%.

\subsection{Synaptic plasticity}

Initial synaptic weights were all equal. Their value was computed so that $\overline{V}_{\mathrm{noise}}=\theta+\sigma_{\mathrm{noise}}$ (leading to an initial firing rate of about 4Hz, see Figure~\ref{fig:stdp} top). They then evolved in $[0,1]$ with all-to-all spike STDP. Yet, we only modeled the Long Term Potentiation part of STDP, ignoring its Long Term Depression (LTD) term.  As in~\cite{Song2000}, we used a trace of presynaptic spikes at each synapse $i$, $A_{\mathrm{pre}}^i$, which was incremented by $\delta A_{\mathrm{pre}}$ at each presynaptic spike, and then exponentially decayed towards 0 with a time constant $\tau_{\mathrm{pre}}=20$ms. At each postsynaptic  spike this trace was used for LTP at each synapse: $w_i\rightarrow w_i+w_i(1-w_i)A_{\mathrm{pre}}^i$.

Here LTD was modeled by a simple homeostatic mechanism. At each postsynaptic spike, all synapses were depressed:  $w_i\rightarrow w_i+w_i(1-w_i)w^{\mathrm{out}}$
where $w^{\mathrm{out}}<0$ is a fixed parameter~\citep{Kempter1999}.

Note that for both LTP and LTD we used the multiplicative term $w_i(1-w_i)$, in contrast with additive STDP, with which the $\Delta w$ is independent of the current weight value~\citep{Song2000,Kempter1999}. This multiplicative term ensures that the weights remain in the range [0,1], and the weight dependence creates a soft bound effect: when a weight approaches a bound, weight changes tend toward zero.  Here it was found to increase performance (convergence time and stability), in line with our previous studies~\citep{Masquelier2007,Kheradpisheh2015,Mozafari2017,Kheradpisheh2018}.

The ratio between LTP and LTD, that is between $\delta A_{\mathrm{pre}}$ and $w^{\mathrm{out}}$ is crucial: the higher, the more synapses are maximally potentiated ($w=1$) after convergence. Here we chose to keep $\delta A_{\mathrm{pre}}=0.1$ and to systematically vary $w^{\mathrm{out}}$, using again a geometric progression with a ratio of 2.5\%. 

\begin{figure}[!htb]
	\centering
	\includegraphics[width=0.9\linewidth]{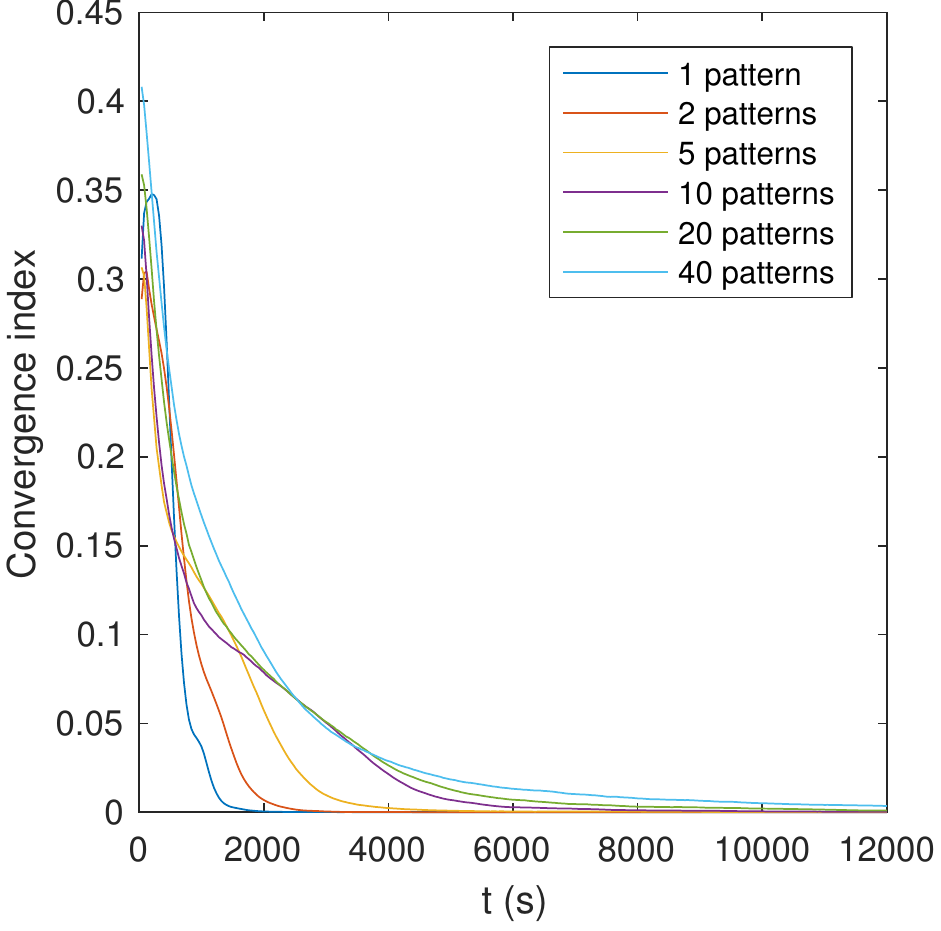}
	\caption{Convergence index as a function of time and number of patterns, for an example of optimal simulation. The convergence index is defined as the mean distance between the full precision weights, and their binary quantization (0 if $w<0.5$, and 1 otherwise).}
	\label{fig:conv}
\end{figure}

\subsection{Results}

For each $\theta_0 \times w^{\mathrm{out}}$ point, 100 simulations were performed with different random pattern realizations, and we computed the proportion of ``optimal'' ones (see below), and reported it in Table~\ref{tab:perf}. After 12,000s of simulated time, the synaptic weights had all converged by saturation. That is synapses were either completely depressed ($w=0$), or maximally potentiated ($w=1$). A simulation was considered optimal if

\begin{enumerate}
	\item \emph{all} the patterns were learned, and
	\item in an optimal way, that is if all patterns exhibited a subsection in which all spikes corresponded to maximally potentiated synapses ($w=1$), and whose duration roughly matched the theoretical optimal $\Delta t$. In practice, we used the total number of potentiated synapses as a proxy of the mean subsection duration (since there is a non-ambiguous mapping between the two variables, given by Equation~\ref{eq:M}), and checked if this number matched the theoretical optimal $M$ (Eq.~\ref{eq:M}) with a 5\% margin.
\end{enumerate}

Note that this second condition alone would be easy to satisfy: the total amount of potentiated synapses is determined by the LTP/LTD ratio which we adjusted by fine-tuning $w^{\mathrm{out}}$. However, satisfying the two conditions is harder, especially when $P$ increases (Table~\ref{tab:perf}).

It is worth mentioning that the learned subsections always corresponded to the beginning of the patterns, because STDP tracks back through them~\citep{Masquelier2008,Masquelier2009,Gilson2011}, but this is irrelevant here since all the subsections are equivalent for the theory. Figure~\ref{fig:stdp} shows an optimal simulation with $P=2$ patterns.

As can be seen in Table~\ref{tab:perf}, the proportion of optimal simulations decreases with $P$, as expected. But more surprisingly, several tens of patterns can be optimally learned with reasonably high probability. With $P=40$ the probability of optimal simulations is only 58\%, but the average number of learned patterns is high: 39.5! This means that nearly all patterns are learned in all simulations, yet sometimes in a suboptimal manner. Finally, Figure~\ref{fig:conv} shows that convergence time increases with $P$.

\begin{table*}[!htb]
	\centering
	\caption{Performance as a function of the number of patterns $P$. The first four lines are computed from the theoretical optimum. The next two lines are the optimal values found through exhaustive search (see text). The last four lines are performance indicators, estimated during the last 100 presentations of each pattern. $\left<P_{\mathrm{learned}}\right>$ is the mean number of ``learned patterns'', that is by convention patterns which elicit at least one postsynaptic spike. The following line is the mean hit rate for those patterns. The subsequent line gives the false alarm rate, but we never observed any here. Finally P(opt) is the proportion of optimal cases.}
	\vspace{0.25cm}
	\begin{tabular}{l l l l l l l}
		\hline
		$P$ & 5 & 10 & 20 & 40 \\
		\hline
		$\Delta t^{\mathrm{opt}}$ (ms) & 11 & 8.1 & 5.7 & 3.7 \\
		$\tau^{\mathrm{opt}}$ (ms) & 8.9 & 6.8 & 5.6 & 5.1 \\
		$M^{\mathrm{opt}}$  & 1600 & 2300 & 3100 & 3800 \\
		$SNR^{\mathrm{opt}}$  & 31 & 20 & 12 & 6.7 \\
		\hline
		$\theta_0$  & 190 & 140 & 110 & 92 \\
		$w^{\mathrm{out}}$  & $-6.2\,10^{-3}$ & $-6.3\,10^{-3}$ & $-6.5\,10^{-3}$ & $-6.7\,10^{-3}$ \\
		\hline
		$\left<P_{\mathrm{learned}}\right>$ & 5 & 10 & 20 & 39.5 \\
		Hit rate (\%)  & 98.9 & 98.6 & 97.9 & 96.5 \\
		False alarms (Hz)  & 0 & 0 & 0 & 0 \\
		P(opt) (\%)  & 100 & 100 & 100 & 58 \\
	\end{tabular}
	\label{tab:perf}
\end{table*}

\section{Discussion}

The fact that STDP can generate selectivity to any repeating spike pattern in an unsupervised manner is a remarkable, yet well documented fact~\citep{Masquelier2008,Masquelier2009,Gilson2011,Humble2012,Hunzinger2012,Klampfl2013,Nessler2013,Kasabov2013a,Krunglevicius2015a,Sun2016,Yger2015,Masquelier2016a}. Here we have shown that, surprisingly, a single neuron can become optimally selective to several tens of independent patterns. Hence STDP and coincidence detection are compatible with distributed coding.

Yet one issue with having one neuron selective to multiple patterns is stability. If one of the learned pattern does not occur for a long period during which the other patterns occur many times, causing postsynaptic spikes, the unseen pattern will tend to be forgotten. This is not an issue with localist coding: if the learned pattern does not occur, the threshold is hardly ever reached so the weights are not modified, and the pattern is retained indefinitely, even if STDP is ``on'' all the time.

Another issue with distributed coding is how the readout is done, that is how the identity of the stimulus can be inferred from multiple neuron responses, given that each response is ambiguous? This is out of the scope of the current paper, but we suspect that STDP could again help. As shown in this study, each neuron equipped with STDP can learn to fire to multiple independent stimuli. Let's suppose that stimuli are shown one at a time. When stimulus $A$ is shown, all the neurons that learned this stimulus (among others) will fire synchronously. Let us call $S$ this set of neurons. A downstream neuron equipped with STDP could easily become selective to this synchronous volley of spikes from neurons in $S$~\citep{Brette2012}. With an appropriate threshold, this neuron would  fire if and only if all the neurons in $S$ have fired. Does that necessarily mean that A is there? Yes, if the intersection of the sets of stimuli learned by neurons in $S$ only contains $A$. In the general case, the intersection is likely to be much smaller than the typical sets of stimuli learned by the $S$ neurons, so much of the ambiguity should be resolved.

What could determine the set of patterns to which a neuron responds? Here, we used independent, unrelated, patterns (i.e. with chance-level overlap), and yet several of these patterns could be learned by a single neuron. Of course, patterns with more overlap would be easier to group. So in the presence of multiple postsynaptic neurons, each one would tend to learn a cluster of similar patterns. Another factor is the time at which the patterns are presented: those presented at the same period are more likely to be learned by the same neuron -- a neuron which was still unselective at that period. Indeed, neurons equipped with STDP have some sort of critical period, before convergence, during which they can learn new pattern easily. Conversely, after convergence, neurons tend to fire if and only if the patterns they have learned are presented (Fig.\ref{fig:stdp}), and thus can hardly learn any new pattern. This is interesting, because patterns presented at the same period are likely to be somewhat related. For example, a neuron could fire to the different people you have met on your first day at work. In the presence of neurogenesis, newborn neurons could handle the learning of other patterns during the subsequent periods of your life. Finally, here we did not use any reward signal. But such a signal, if available, could modulate STDP (leading to some form of supervised learning), and encourage a given neuron to fire to a particular, meaningful, set of patterns~\citep{Mozafari2018a,Mozafari2018}, as opposed to a random set like here. For e.g. a single neuron could learn to fire to any animal, even if different animals cause very different sensory inputs.

Here the STDP rule we used always led to binary weights after learning. That is an afferent could be either selected or discarded. We thus could use our $SNR$ calculations derived with binary weights, and checked that the selected set was optimal given the binary weight constraint. Further calculations in the Appendix suggest that removing such a constraint could lead to a modest increase in $SNR$, of about 10\%. More research is needed to see if a multiplicative STDP rule, which does not converge towards binary weights~\citep{VanRossum2000,Gutig2003}, could lead to the optimal graded weights.

Our theoretical study suggests, together with others~\citep{Gutig2006,Brette2012}, that coincidence detection is computationally powerful. In fact, it could be the main function of neurons~\citep{Abeles1982,Konig1996}. In line with this proposal, neurons \emph{in vivo} appear to be mainly fluctuation-driven, not mean-driven~\citep{Rossant2011,Brette2012,Brette2015}. This is the case in particular in the balanced regime~\citep{Brette2015}, which appears to be the prevalent regime in the brain~\citep{Deneve2016}. Several other points suggest that coincidence detection is the main function of neurons. Firstly, strong feedforward inhibitory circuits throughout the central nervous system often shorten the neurons' effective integration windows~\citep{Bruno2011}. Secondly, the effective integration time constant in dendrites might be one order of magnitude shorter than the soma's one~\citep{Konig1996}. Finally, recent experiments indicate that a neuron's threshold quickly adapts to recent potential values~\citep{Platkiewicz2011,Fontaine2014,Mensi2016}, so that only a sudden potential increase can trigger a postsynaptic spike. This enhances coincidence detection. It remains unclear if other spike time aspects such as ranks~\citep{Thorpe1998} also matter.

Our results show that lower firing rates lead to better signal-to-ratio. It is worth mentioning that mean firing rates are probably largely overestimated in the electrophysiological literature, because extracellular recordings -- by far the most popular technique -- are totally blind to cells that do not fire at all~\citep{Thorpe2011}. Even a cell that fire only a handful of spikes will be ignored, because spike sorting  algorithms need tens of spikes from a given cell before they can create a new cluster corresponding to that cell. Furthermore, experimentalists tend to search for stimuli that elicit strong responses, and, when they can move the electrode(s), tend to look for most responsive cells, introducing strong selection biases. Mean firing rates, averaged across time and cells, are largely unknown, but they could be smaller than 1 Hz~\citep{Shoham2006}. It seems like coding is sparse: neurons only fire when they need to signal an important event, and that every spike matters~\citep{Wolfe2010}.

Finally, we see an analogy between our theory, and the one of  neural associative memory (NAM), in which an output (data) vector is produced by multiplying an input (address) vector by a weight matrix. Unlike NAM, our framework is dynamic, yet after learning, to a first approximation, our STDP neurons count the number of input spikes arriving through reinforced synapses in a short integration window, and each one outputs a 1 (i.e. a spike) if this count exceeds a threshold, and a 0 otherwise, leading to a binary output vector, much like in a binary NAM. It is thus unsurprising that  sparsity is desirable both in our theory, and in NAMs~\citep{Palm2013}.

\subsection*{Acknowledgments}
This research received funding from the European Research Council under the European Union's $7^{\mathrm{th}}$
Framework Program (FP/2007-2013) / ERC Grant Agreement n.323711 (M4 project).
We thank Milad Mozafari for smart implementation hints, and Jean Pierre Jaffr\'ezou for his excellent copy editing.

\section*{Appendix}
\subsection*{Graded weights}
\label{sec:graded}

In this paper, we assumed unitary (or binary) synaptic weights: all connected afferents had the same synaptic weight\footnote{\normalsize Numerical simulations with STDP used graded weights during learning, but not after convergence.}. This constraint strongly simplified the analytical calculations. But could the $SNR$ be even higher if we removed this constraint, and by how much? Intuitively, when one wants to detect a spike pattern that has just occurred, one should put strong weights on the synapses corresponding to the most recent pattern spikes, since these weights will increase $V_{\max}$ more than $V_\mathrm{noise}$. Conversely, very old pattern spikes that fall outside the integration window (if any) should be associated to nil weights: any positive value would only increase $V_\mathrm{noise}$, not $V_{\max}$. But between those two extremes, it might be a good idea to use intermediate weight values.

\begin{figure}[!htb]
	\centering\includegraphics[width=1.0\linewidth]{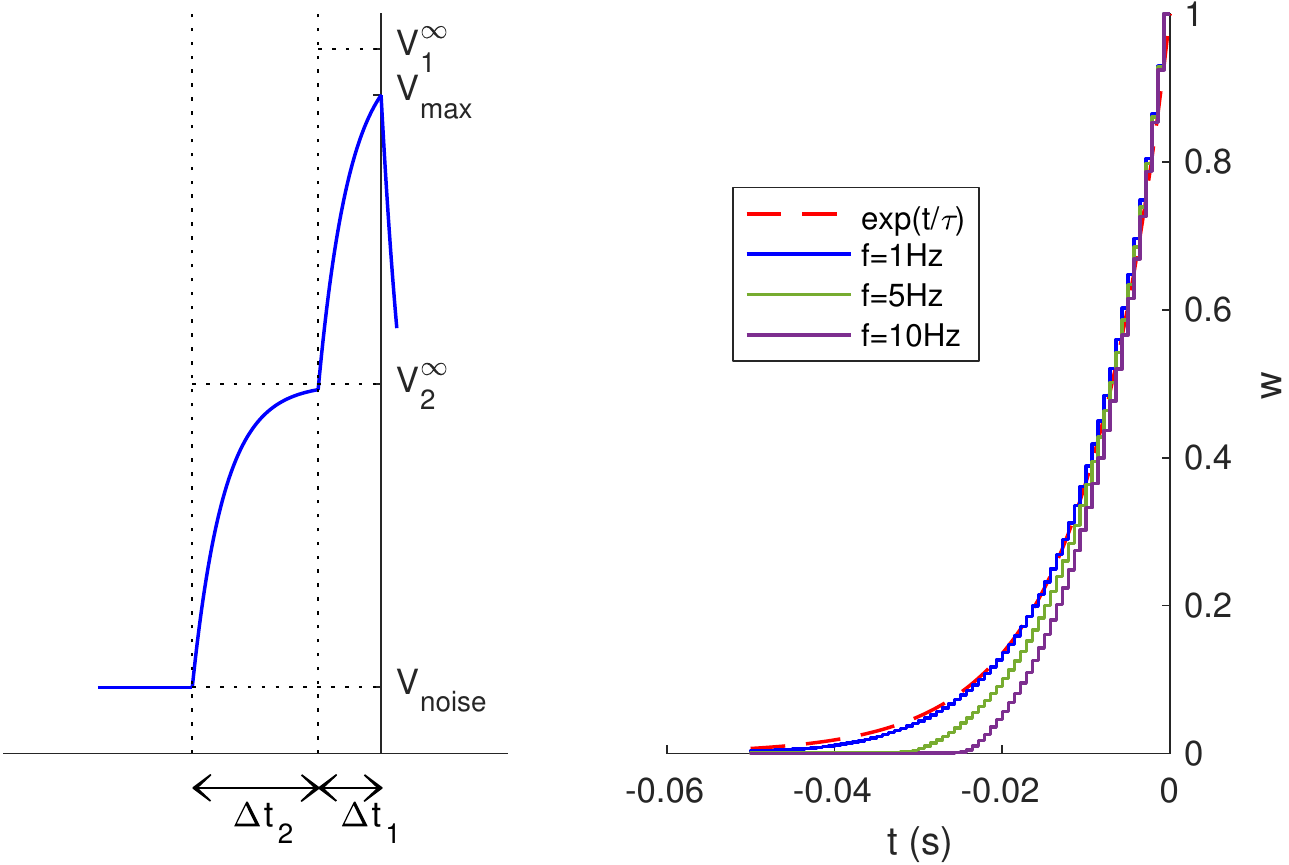}
	\caption{Optimization with graded weights. (Left) Didactic example with $n=2$ weight values: $w_1$ for all the afferents that fire in the $\Delta t_1$ window, and $w_2 < w_1$ for all the afferents that fire in the $\Delta t_2$ window but not in the $\Delta t_1$ one. $V_1^\infty$ and $V_2^\infty$ are the asymptotic potentials for the two periods. $V_{\max}$ can be computed from those two values (see text). (Right) Numerical optimization of the weights with $n=70$. With small $f$, the optimal solution appears to be close to $e^{t/\tau}$.
	}
	\label{fig:graded}
\end{figure}

To check this intuition, we used numerical optimizations using a simplified setup. We used a single pattern ($P=1$), that was repeated in the absence of jitter ($T=0$). We divided the pattern into $n$ different periods $\Delta t_1, ...  \Delta t_n$ (in reverse chronological order), each one corresponding to a different synaptic weight $w_1, ... w_n$ (see Figure~\ref{fig:graded} left for an example with $n=2$). More specifically: the $M_1$ afferents that fire in the $\Delta t_1$ window are connected with weight $w_1$. The $M_2$ afferents that fire in the $\Delta t_2$ window, but not in the $\Delta t_1$ one, are connected with weight $w_2$. More generally, the $M_i$ afferents that fire in the $\Delta t_i$ window, but not in the $\Delta t_1 ... \Delta t_{i-1}$ ones, are connected with weight $w_i$.
 
With this simple set up, the $SNR$ can be computed analytically. For example, if $n=2$ (Fig.~\ref{fig:graded} left), we have:
\begin{equation}
\label{eq:graded_begin}
\left<M_1\right>=N(1-e^{-f\Delta t_1}),
\end{equation}
\begin{equation}
\left<M_2\right>=N(1-e^{-f\Delta t_2})e^{-f\Delta t_1}.
\end{equation}
The asymptotic steady regimes for the two time windows are:
\begin{equation}
\left<V_1^\infty\right>=\tau f w_1 N,
\end{equation}
\begin{equation}
\left<V_2^\infty\right>= \tau f \left( w_2 N + (w_1 - w_2) \left<M_1\right> \right).
\end{equation}
Let's call $V_i$ the potential at the end of window $\Delta t_i$, and $V_{n+1}=V_\mathrm{noise}$. Then $V_{\max}=V_1$ can be computed iteratively:
\begin{equation}
V_2 = (1-e^{-\Delta t_2 / \tau})(V_2^\infty - V_3),
\end{equation}
\begin{equation}
V_1 = (1-e^{-\Delta t_1 / \tau})(V_1^\infty - V_2).
\end{equation}
Furthermore~\citep{Burkitt2006a},
\begin{equation}
V_\mathrm{noise} = \tau f (w_1 M_1 + w_2 M_2),
\end{equation}
and:
\begin{equation}
\label{eq:graded_end}
\sigma_\mathrm{noise}  = \sqrt{\tau f (w_1^2 M_1 + w_2^2 M_2)/2}.
\end{equation}
So we have everything we need to compute the $SNR$.

Equations~\ref{eq:graded_begin}~--~\ref{eq:graded_end} can be generalized to $n>2$:
\begin{equation}
\left<M_i\right>=N(1-e^{-f\Delta t_i})e^{-f\sum\limits_{j=1}^{i-1}\Delta t_j	},
\end{equation}
\begin{equation}
\left<V_i^\infty\right>= \tau f \left( w_i N + \sum\limits_{j=1}^{i-1} (w_j - w_i) \left<M_j\right> \right)
\end{equation}
and $V_{\max}=V_1$ can be computed iteratively from $V_{n+1}=V_\mathrm{noise}$ using:
\begin{equation}
V_{i-1} = (1-e^{-\Delta t_{i-1} / \tau})(V_{i-1}^\infty - V_i).
\end{equation}
Furthermore~\citep{Burkitt2006a},
\begin{equation}
V_\mathrm{noise} = \tau f \sum w_i M_i,
\end{equation}
and:
\begin{equation}
\sigma_\mathrm{noise}  = \sqrt{\tau f  \sum w_i^2 M_i/2}.
\end{equation}

So the $SNR$ can be computed for any $n$, and, importantly, it is differentiable with respect to the $w_i$. We can thus use efficient numerical methods to optimize these weights. Since scaling the weights does not change the $SNR$, we imposed $w_1=1$. Figure~\ref{fig:graded} right gives an example with $n=70$. Here the $\Delta t_i$ were all equal to $5\tau/n$, and we optimized the corresponding $w_i$. We chose $\tau=10$ms, and $f=1$, 5, and 10Hz. The gain w.r.t. binary weights for the $SNR$ were modest: 10.5\%, 9.6\% and 8.9\% respectively. As  $f$ tends towards 0, the optimal weights appears to converge towards $e^{t/\tau}$ (even if we could not prove it): the $f=1$Hz curve (solid blue) is almost identical to $e^{t/\tau}$ (dashed red).

%\bibliographystyle{apa}
%\bibliography{../library}

\end{document}